\documentclass[letterpaper, 10 pt, conference]{ieeeconf}  
\usepackage{amsmath}
\usepackage{algorithm}
\usepackage[noend]{algpseudocode}
\usepackage{stfloats} 
\IEEEoverridecommandlockouts                              
\usepackage{tabularx}
\usepackage{makecell}
\usepackage{pifont}                                                     
\usepackage{placeins}
\usepackage{amsfonts}
\usepackage{graphicx}
\usepackage{subfig}
\usepackage{rotating}
\usepackage{booktabs} 
\graphicspath{{figures/}}
\usepackage{multirow}
\usepackage{hhline}
\usepackage{amsmath}
\usepackage{adjustbox}
\usepackage{caption}
\usepackage{fancyhdr}
\usepackage{lipsum}  
\usepackage{algorithm}
\usepackage[noEnd=false,indLines=false]{algpseudocodex}

\newcommand{\Input}{\item[\textbf{Input:}]}

\renewcommand{\Return}{\State \textbf{return~}}
\usepackage{booktabs} 
\usepackage{amsmath}
\usepackage{calrsfs}

\newcommand{\RNum}[1]{\lowercase\expandafter{\romannumeral #1\relax}}
\usepackage[font=small]{caption}

\title
{\LARGE \bf
 A Multi-Modal Interaction Framework for Efficient Human-Robot Collaborative Shelf Picking}
\author{Abhinav Pathak$^{1,2,\dagger}$, Kalaichelvi Venkatesan$^{2}$, Tarek Taha$^{1}$ and Rajkumar Muthusamy$^{1,\dagger,\ast}$ \thanks{$^{\dagger}$Equal contribution. $^{\ast}$Corresponding Author.} 
\thanks{$^{1}$Robotics Lab, Dubai Future Labs, Dubai, UAE. Email: \tt\small{rajkumar.muthusamy@dubaifuture.gov.ae}} 
\thanks{$^{2}$Birla Institute of Technology and Science, Pilani, Dubai Campus, UAE} 
}
\begin{document}
\maketitle
\thispagestyle{empty}
\pagestyle{empty}

\begin{abstract}
The growing presence of service robots in human-centric environments, such as warehouses, demands seamless and intuitive human-robot collaboration. In this paper, we propose a collaborative shelf-picking framework that combines multimodal interaction, physics-based reasoning, and task division for enhanced human-robot teamwork. 

The framework enables the robot to recognize human pointing gestures, interpret verbal cues and voice commands, and communicate through visual and auditory feedback. Moreover, it is powered by a Large Language Model (LLM) which utilizes Chain of Thought (CoT) and a physics-based simulation engine for safely retrieving cluttered stacks of boxes on shelves, relationship graph for sub-task generation, extraction sequence planning and decision making. Furthermore, we validate the framework through real-world shelf picking experiments such as 1) Gesture-Guided Box Extraction, 2) Collaborative Shelf Clearing and 3) Collaborative Stability Assistance. 
\end{abstract}
\IEEEpeerreviewmaketitle
\section{Introduction} \label{intro}

The rapid integration of robotic systems into human-centric environments, such as warehouses and logistics centers, has fundamentally altered the landscape of Human-Robot Collaboration (HRC). While robots bring unparalleled precision, endurance, and strength, humans contribute indispensable dexterity, adaptability, and decision-making capabilities. However, achieving seamless and intuitive collaboration remains a significant hurdle. Current robotic systems often operate within rigid frameworks, demanding precise, pre-defined instructions that hinder natural interaction. This disconnect creates a substantial barrier to effective teamwork, especially in dynamic and unstructured environments where human intuition and adaptability are paramount. Specifically, robots struggle to interpret the nuances of human intent, requiring specialized commands and limiting natural communication.

Beyond the challenge of interpreting human intent, robotic systems also exhibit a significant deficiency in understanding their physical surroundings. This lack of physical reasoning manifests in an inability to accurately predict how objects interact, particularly in complex scenarios like shelf picking. For instance, robots often fail to recognize the intricate relationships between stacked boxes, leading to potential instability and collapse when attempting to remove specific items. This failure to comprehend the physical dynamics of their environment underscores a critical gap in their ability to operate safely and effectively in cluttered spaces.

\begin{figure}
    \centering
    \includegraphics[width=0.85\linewidth]{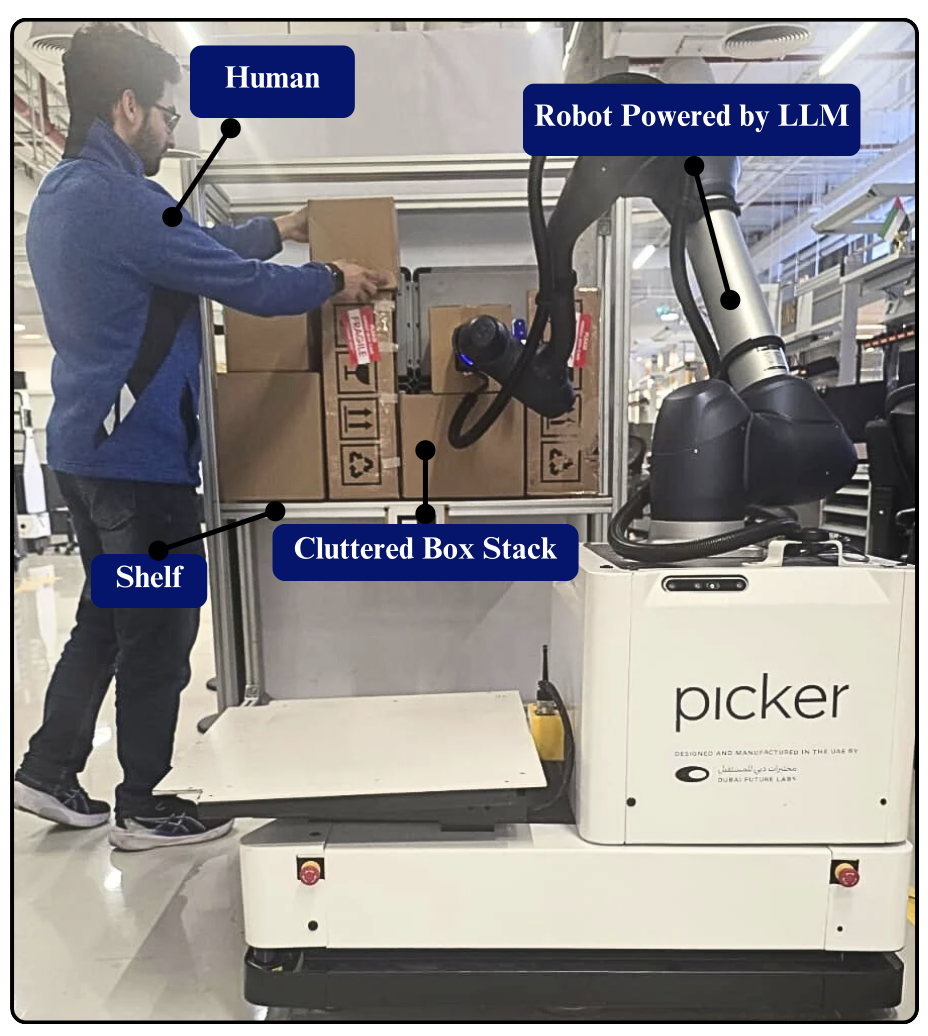}
    \caption{Collaborative Shelf Picking: This illustration depicts a mobile manipulator robot powered by a LLM and a novel physics-based reasoning engine collaborating with the human in real-time}
    \label{fig:enter-label}
    \vspace{-1.5em}
\end{figure}

To overcome the limitations of current robotic systems and enable truly collaborative human-robot interaction, we introduce a new collaborative shelf-picking framework that goes beyond present robotic technology to support interaction between humans and robots. The framework can handle multi-modal inputs consisting of verbal commands, natural language and gestures and can also communicate back via auditory and visual feedback. Additionally, it is powered by a LLM with chain-of-thought reasoning along with a novel physics reasoning engine that enables it to plan a safe extraction sequence that prevents box collapse and collisions. Building upon our prior work \cite{pathakpaper} on real-to-simulation pipelines for extraction sequence planning, we extend the concept from physics-based reasoning to further analyze box relationships and support structures, enhancing safety and stability. The system delivers proactive assistance, dynamic task allocation, and transparent information exchange, resulting in safer and more efficient operations.

Key contributions include:

\begin{enumerate}
\item Multimodal Integration: A collaborative shelf-picking framework that integrates multimodal interactions (gesture, natural language) and provides real-time auditory and visual feedback, powered by an LLM with Chain-of-Thought (CoT) and physics-informed reasoning, enabling intuitive human-robot alignment.
\item Physical Reasoning: A Physics-informed Box Relations Graph (BRG) derived from RGB-D data, coupled with physics simulation, to accurately model structural dependencies and predict potential box collapse during shelf-picking, ensuring safe and stable operation.
\item Task Allocation: BRG-driven dynamic task partitioning and assistance, allowing for adaptive role allocation and proactive support based on structural dependencies, enhancing collaborative efficiency.
\item Experimental Validation: Validation of the framework through real-world shelf-picking experiments, including (a) Gesture-Guided Box Extraction, (b) Collaborative Shelf Clearing, and (c) Collaborative Stability Assistance, demonstrating its efficacy in complex scenarios.
\end{enumerate}
The paper is structured as follows: Section \ref{Related Work} reviews related work in object retrieval, simulation-based planning, and physics-informed robotics. Section \ref{methodology} details the proposed methodology, including the perception pipeline, simulation techniques, and decision-making framework. Experimental results are presented in Section \ref{exp}, comparing the performance of the proposed approach in both simulation and real-world scenarios. Finally, Section \ref{conc} provides conclusions, discusses limitations, and outlines potential future research directions.

\begin{figure*}[th!]
    \centering
    \includegraphics[width=0.85\linewidth]{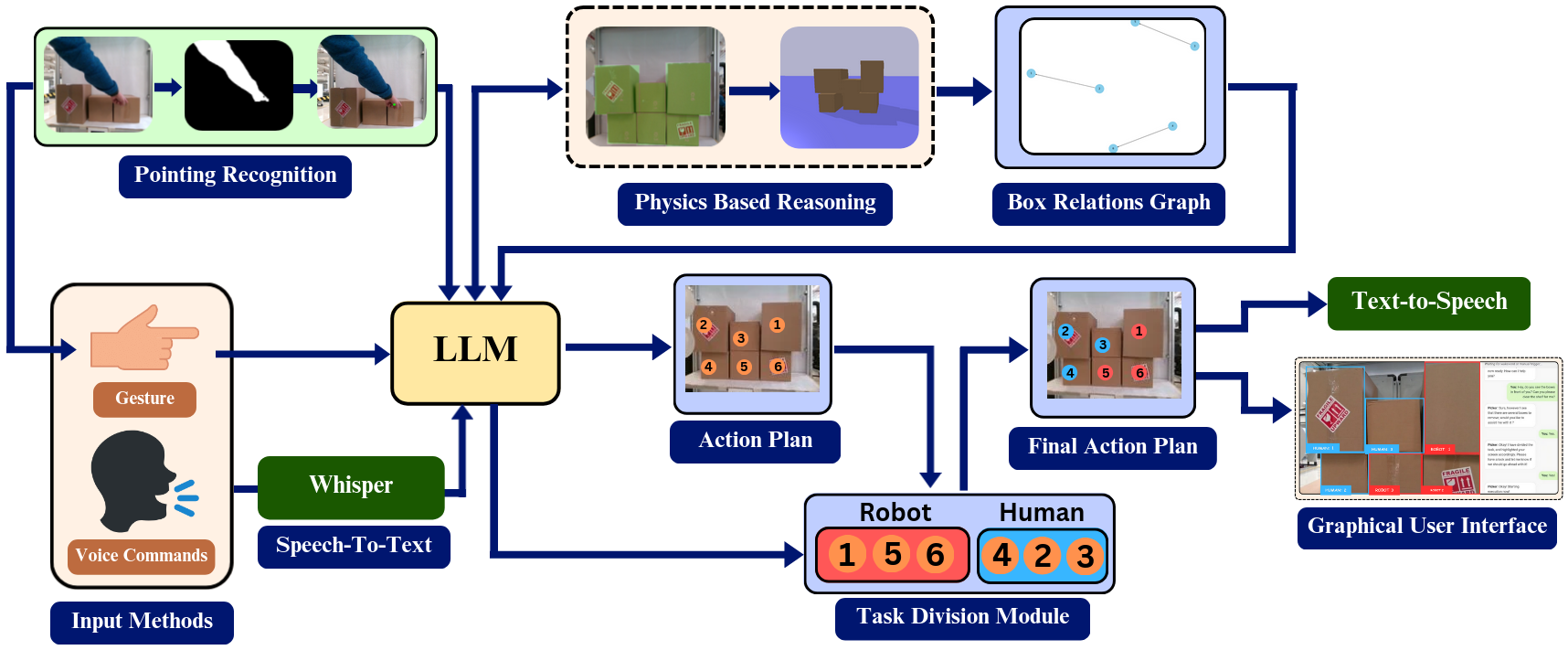}
    \caption{Overview of the proposed grasping pipeline using the physics-aware approach for safe cardboard box extraction}
    \label{fig:graspplanner}
    \vspace{-1.5em}
\end{figure*}

\section{Related Work} \label{Related Work}

Human–robot collaboration (HRC) in warehouse environments has gained significant attention due to its potential to enhance efficiency, safety, and adaptability in dynamic settings. Research in this field spans various aspects, including task optimization, anticipatory planning, and simulation-based reasoning.

Optimizing order picking is a key challenge in warehouse automation. Zhao et al. \cite{10478104} present an integrated framework that jointly addresses pod selection, robot scheduling, and manual picking, highlighting the importance of coordination between human workers and robotic systems. Empirical studies by Pasparakis et al. \cite{Pasparakis2021InCO} and Jacob et al. \cite{JACOB2023109262} further demonstrate that well-designed HRC strategies not only improve productivity but also enhance safety in cluttered warehouse environments.

A major challenge in shelf picking is the safe extraction of objects from unstable stacks. To mitigate risks, Motoda et al. \cite{9551507} propose a bimanual manipulation planner based on collapse prediction, ensuring structural stability during retrieval. Their subsequent work on multi-step extraction \cite{motoda2023multistep} leverages object support relations to maintain pile integrity, while their shelf replenishment strategy \cite{robotics11050104} integrates object arrangement detection with collapse risk prediction. These works underscore the necessity of predictive reasoning in robotic extraction.

Advancements in vision-based and simulation-driven approaches have significantly improved robotic perception and action planning. Chen et al. \cite{machines11020275} integrate deep reinforcement learning with computer vision for robust grasping, while Bejjani et al. \cite{bejjani2021occlusionawaresearchobjectretrieval} tackle the challenge of retrieving occluded objects from clutter. Complementarily, Zook et al. \cite{zook2024grsgeneratingroboticsimulation} introduce a real-to-sim pipeline that converts single RGB-D observations into digital twin environments, providing robots with simulation spaces for training reinforcement learning models. Ni and Qureshi \cite{ni2023progressivelearningphysicsinformedneural} further refine motion planning by incorporating physics-informed neural networks, improving collision avoidance and path efficiency.

In parallel, the integration of natural language interfaces is emerging as a promising direction in HRC. Long et al. \cite{long2024robollmroboticvisiontasks} demonstrate that multi-modal large language models can effectively bridge robotic vision and language understanding, reducing the need for complex task-specific encoders. However, existing models remain unreliable for real-world deployment, and their reliance on cloud-based inference presents latency and scalability challenges.

Foundational surveys provide additional insights relevant to our work. Bohg et al. \cite{graspplanning} offer a comprehensive review of data-driven grasp synthesis methods, while Banerjee et al. \cite{banerjee2024physicsinformedcomputervisionreview} explore the integration of physical laws into computer vision, reinforcing the importance of embedding physics-based reasoning into robotic perception. Additionally, trust and transparency play a crucial role in effective HRC. Chen and Chan \cite{10569153} review trust-aware shared control mechanisms, while Khanna et al. \cite{khanna2023userstudyexploringrole} examine how explaining robotic failures improves user trust and collaboration.

Current HRC research struggles with real-world shelf-picking complexities. We address this by integrating a real-to-sim pipeline with an LLM interface. This enables robots to interpret natural language, understand human intent, assess extraction, and predict risks like collapse. This integration fosters efficient, safe, and adaptive collaboration in diverse warehouse environments
\section{Methodology} \label{methodology}

This section details a collaborative shelf picking framework that is capable of multi-modal interactivity to enable more effective human robot collaboration. The framework leverages a Large Language Model (LLM) with Chain of Thought (CoT), integrated with a novel physics simulation engine. This engine models structural support and potential instabilities, grounding the LLM's reasoning in real-world physics. Additionally, the framework incorporates visual perception and multi-modal interaction—natural language commands and gesture recognition—to create shared situational awareness. These components enable the robot to perceive and understand its environment, improving planning, reducing risk, and increasing collaborative efficiency. The following subsections provide a detailed breakdown of each component within the proposed framework.
\subsection{Scene Perception}

Using a single RGB-D image a segmenetation mask for all the detected cardboard boxes in the scene is generated. From this segmentation mask, box dimensions, location (centroid), orientation, and distance from the camera are calculated using bounding boxes and depth data.
\begin{figure}[H]
    \centering
    \includegraphics[width=0.85\linewidth]{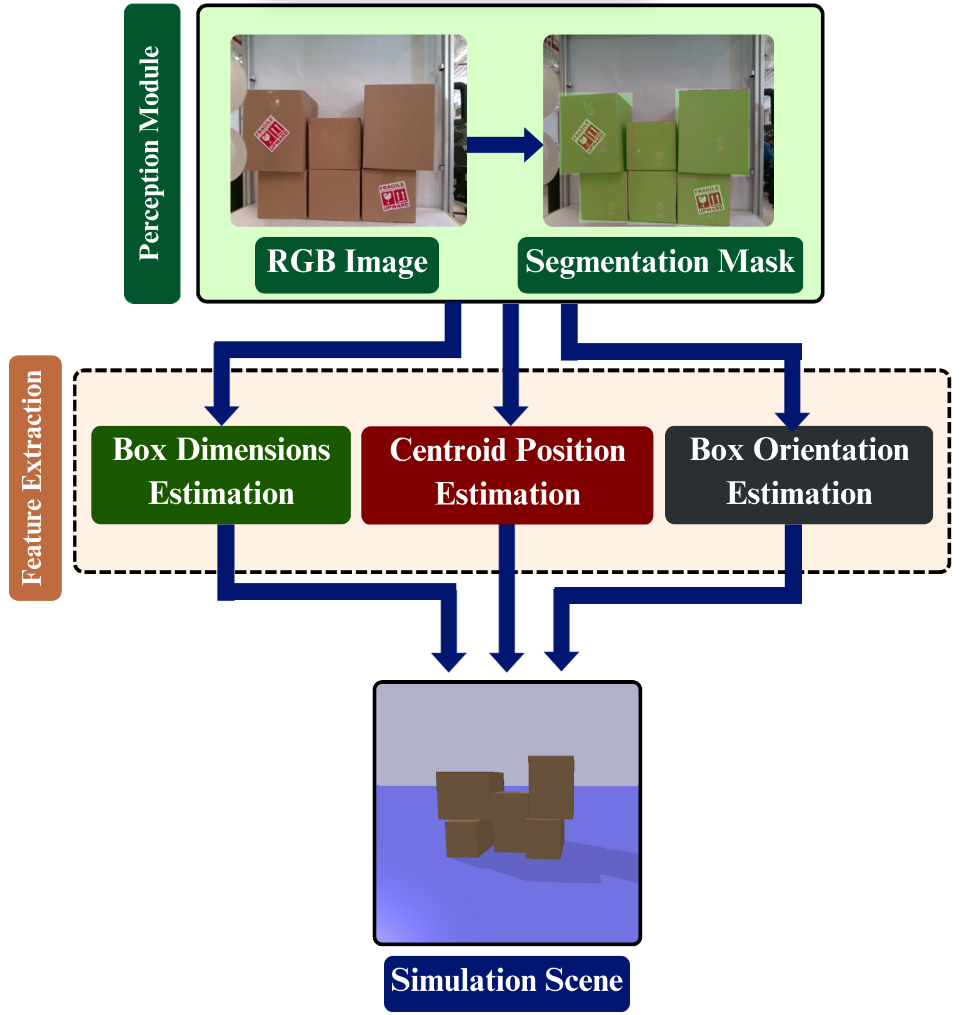}
    \caption{Real-to-Sim Pipeline}
    \label{fig:real-to-sim}
    \vspace{-1.5em}
\end{figure}

These extracted features are used to reconstruct the perceived scene in a physics simulation using PyBullet, a lightweight and realistic physics engine. Gravity is set to 9.81 m/s², and each box is modeled with a density of 1 kg/m³ and uniform mass distribution. The simulation standardizes interactions by setting the surface friction coefficient to 0.75 and the spinning friction to 0.01 while applying uniform friction across contact surfaces. These parameters provide an accurate model of stacked cardboard structures and yield realistic simulation outcomes. This entire pipeline can be seen in Figure \ref{fig:real-to-sim}.

\subsection{Physics-based Reasoning for Human-Robot Collaboration}\label{ch0-Planning-cp0}



\begin{figure}[H]
    \centering
    \includegraphics[width=1\linewidth]{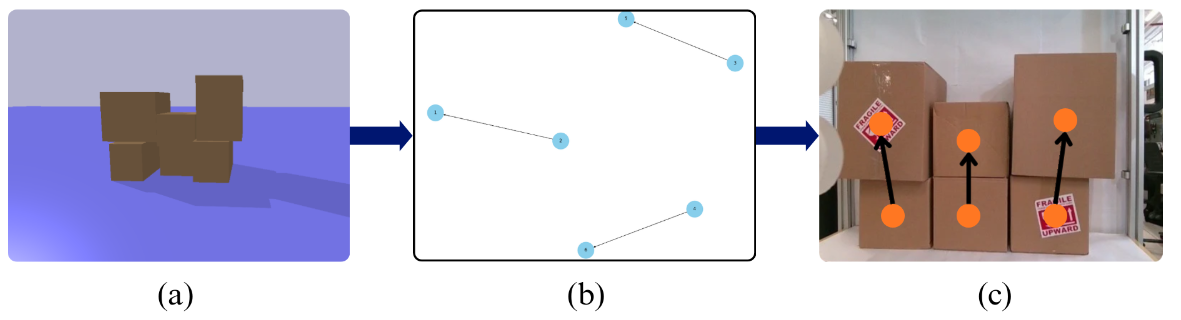}
    \caption{Illustration showing how a  Box Relations Graph is computed from a simulation}
    \label{fig:sim-to-brg}
\end{figure}

Insights from the simulation are used to construct a Box Relations Graph (BRG) (as seen in Figure \ref{fig:sim-to-brg}), which models dependency relationships between boxes. In this graph, each node represents a box and each directed edge indicates a support dependency; an edge is created if removing one box would cause another to collapse. The system employs a depth-first search algorithm to systematically explore these relationships and identify critical structural supports within the stack. This is an extension of our previous work \cite{pathakpaper} as this work focuses more on understanding how the boxes are related to each other in the entire stack, and uses it to compute support candidates and extraction sequences. 

This graph-based approach enables the robot to understand stability constraints and plan box removals in a way that minimizes the risk of unintended collapses, ensuring safe and efficient warehouse operations. Algorithm 1 describes how the BRG is computed.

\begin{algorithm}
\label{algo1}
    \caption{Building the Dependency Graph}
    \begin{algorithmic}
        \Input A dependency dictionary $D$, where for each box $b$, $D[b]$ is a list of boxes that support $b$.
        \State Initialize an empty graph $G$.
        \For{each box $b$ in $D$}
            \State Add node $b$ to $G$.
            \For{each dependency $d$ in $D[b]$}
                \State Add a directed edge from $d$ to $b$ \Comment{Box $b$ depends on $d$.}
            \EndFor
        \EndFor
        \State \Return $G$.
    \end{algorithmic}
\end{algorithm}

\vspace{-1.5em}

Once the Box Relations Graph is constructed, it enables several applications. One of the key applications is safe box extraction (as described in algorithm 2, where Kahn’s algorithm determines an optimal removal sequence for a given target box. The algorithm first identifies all boxes that structurally depend on the target using a recursive traversal of the dependency dictionary. These boxes form a subgraph, which is then topologically sorted to ensure that removal follows dependency constraints. This guarantees that no box is extracted before those it supports, preventing collapses and enabling stable and efficient box removal in automated warehouse systems. Additional applications of the Box Relations Graph are discussed in later sections.

\begin{algorithm}[H]
    \caption{Obtaining the Safe Sequence for Extraction}
    \begin{algorithmic}
        \Input Dependency dictionary $D$, target box $s$.
        \State Initialize an empty set $S_{\text{visited}}$.
        \State Initialize an empty list $S_{\text{sequence}}$.
        \State // Step 1: Recursively collect all boxes related to $s$
        \Function{Explore}{box}
            \If{box $\notin S_{\text{visited}}$}
                \State Add box to $S_{\text{visited}}$.
                \For{each box $b$ in $D[\text{box}]$}
                    \State \Call{Explore}{b}.
                \EndFor
            \EndIf
        \EndFunction
        \State \Call{Explore}{$s$}.
        \State // Step 2: Construct subgraph $G'$ using nodes in $S_{\text{visited}}$ and perform a topological sort
        \State Construct subgraph $G'$ with nodes $S_{\text{visited}}$ and edges from $D$.
        \State Compute a topological ordering of $G'$ and store it in $S_{\text{sequence}}$.
        \State \Return $S_{\text{sequence}}$.
    \end{algorithmic}
\end{algorithm}
\vspace{-1.5em}

\subsection{Interaction and Collaboration}

To enable multi-modal interactions and reasoning, the system consists of the following key components: 

\begin{figure}[h!]
    \centering
    \includegraphics[width=0.8\linewidth]{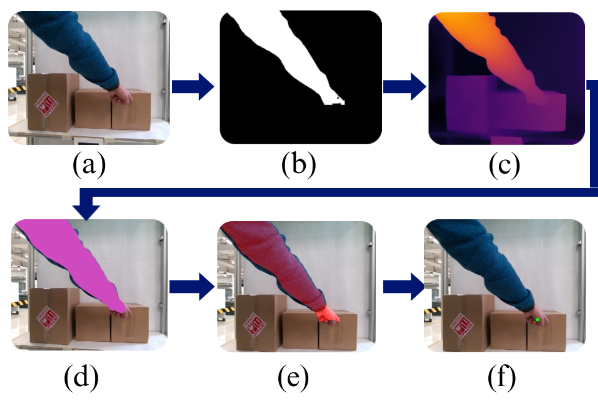}
    \caption{Pointing Recognition Pipeline}
    \label{fig: Pointing Recognition Pipeline}
\end{figure}
\vspace{-1.2em}

\begin{enumerate}
    \item \textbf{Pointing Recognition:}
Captured images are processed using a segmentation mask that isolates the human arm (Figure \ref{fig: Pointing Recognition Pipeline} (b)) . To refine the depth data in Figure \ref{fig: Pointing Recognition Pipeline} (c), the point cloud undergoes DBSCAN clustering, which removes segmentation noise and enhances accuracy (as seen in Figures \ref{fig: Pointing Recognition Pipeline} (d) and (e)). The system then selects the top 2\% of depth points within the refined mask and computes their median to estimate the user's intended target (as seen in Figure \ref{fig: Pointing Recognition Pipeline} (f)). The closest box to this estimated point is identified as the selected box. This enables the LLM to interpret user gestures, providing contextual awareness for decision-making. Based on this, the system can designate the pointed box as the target for extraction. Additionally, it detects whether the user is actively pointing; if a removal request is made without a detected pointing gesture, the system prompts the user to specify the desired box through pointing, ensuring clarity and reducing errors in task execution.

\begin{figure}[H]
    \vspace{-1em}
    \centering
    \includegraphics[width=0.9\linewidth]{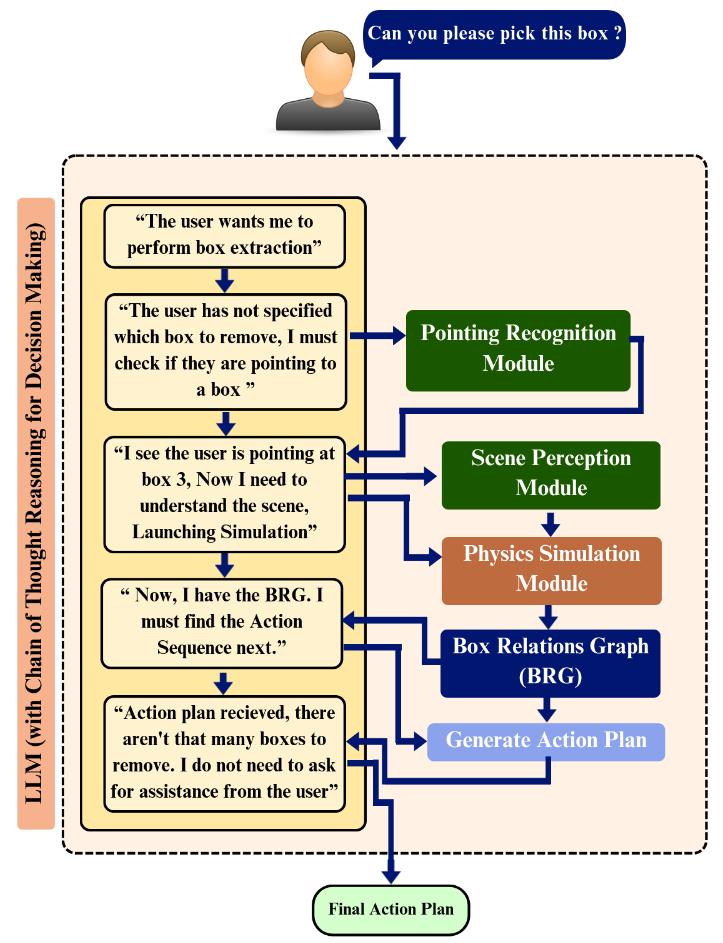}
    \caption{Illustrations showcasing how the LLM uses Chain-of-Thought (CoT) reasoning for decision making}
    \label{fig:LLM Reasoning}

\end{figure}
\item \textbf{Natural Language Reasoning:}
The system also allows users to select a box for extraction through natural language commands. Using the Qwen2.5-7B-Instruct \cite{qwen2.5} model, the extracted features from the perception stage are passed to the LLM, enabling it to exhibit multi-modal capabilities by integrating both visual and textual inputs. With Chain-of-Thought (CoT) prompting, the LLM processes the user’s request, leveraging contextual awareness to accurately identify the referenced box. This ensures a flexible and intuitive interaction, allowing users to specify their intent either through gestures or spoken instructions. The system can also parse natural language queries into sub tasks, which allows it make decisions and give commands to the robot according to the user query. Furthermore, the LLM acts as a bridge between the robot and the human, as it allows the human to natural ask for assistance and give queries without being restricted to a specific querying format or structure. This reasoning is illustrated in Figure \ref{fig:LLM Reasoning}

\item \textbf{Collaboration}

Leveraging the extracted perception features and the Box Relations Graph (BRG) constructed from the physics simulation, the system optimizes efficiency by facilitating human-robot collaboration. It analyzes the predicted action sequence and determines whether task execution can be expedited by prompting human assistance. This can be seen in Figure \ref{fig:llm help}

\begin{figure}[H]
    \centering
    \includegraphics[width=1\linewidth]{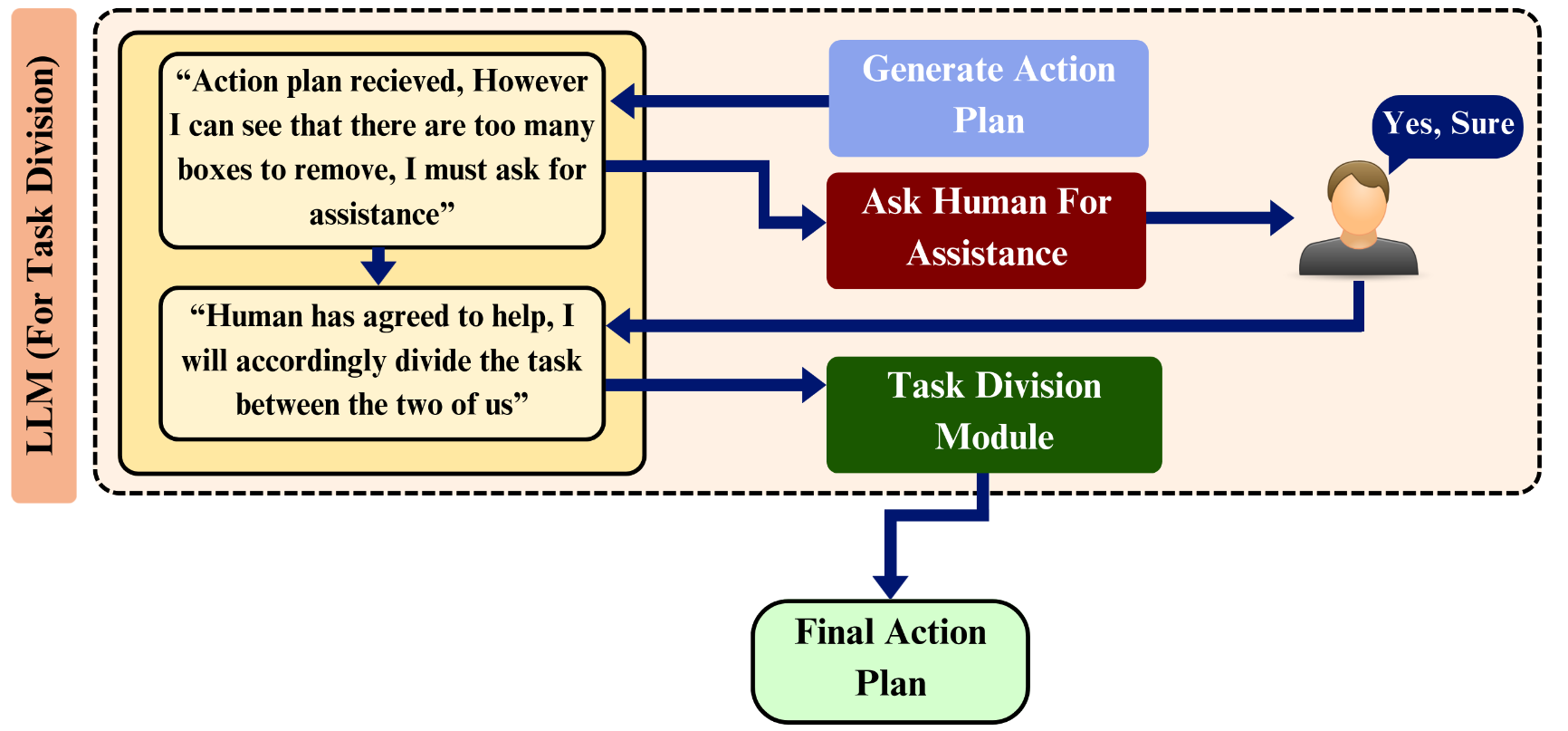}
    \caption{Illustration showcasing how the LLM asks the user for assistance and divides the task accordingly}
    \label{fig:llm help}
    \vspace{-1.5em}
\end{figure}

For shelf clearance, the BRG is used to identify independent nodes—boxes that can be removed without affecting others—allowing the sequence to be split into parallel tasks for both the robot and the human. This is described in algorithm 3.
\begin{algorithm} [H]
    \caption{Dividing the Safe Sequence into Robot and Human Tasks}
    \begin{algorithmic}
        \Input A safe sequence $S$ (an ordered list of boxes) and dependency dictionary $D$.
        \State Initialize empty lists \texttt{RobotTasks} and \texttt{HumanTasks}.
        \For{each box $b$ in $S$}
            \If{$D[b]$ is empty}
                \State Append $b$ to \texttt{RobotTasks} \Comment{Box $b$ has no dependencies.}
            \Else
                \State Append $b$ to \texttt{HumanTasks} \Comment{Box $b$ has dependencies; may require human supervision.}
            \EndIf
        \EndFor
        \State \Return (\texttt{RobotTasks}, \texttt{HumanTasks}).
    \end{algorithmic}
\end{algorithm}

 Similarly, in box extraction, if multiple boxes must be removed before reaching the target, the system can request the user to support certain boxes or directly extract the target while the robot provides the support. By enabling both assistance from and to the human, the system enhances efficiency in shelf-picking tasks through adaptive collaboration. This is described in algorithm 4.

\begin{algorithm}
    \caption{Selecting Support Candidate Boxes}
    \begin{algorithmic}
        \Input Dependency dictionary $D$, target box $s$, maximum number of support candidates $k$.
        \State Initialize an empty set $S_{\text{related}}$.
        \State \Call{Explore}{$s$} \Comment{Reuse the DFS from the safe sequence step to collect related boxes.}
        \State Initialize an empty list \texttt{Candidates}.
        \For{each box $b$ in $S_{\text{related}}$}
            \State Compute the support count: the number of boxes in $D[b]$ that are also in $S_{\text{related}}$.
            \State Append the pair $(b, \text{support count})$ to \texttt{Candidates}.
        \EndFor
        \State Sort \texttt{Candidates} in descending order by support count.
        \State \Return the first $k$ boxes from \texttt{Candidates}.
    \end{algorithmic}
\end{algorithm}

\end{enumerate}

\section{Experiments and Results} \label{exp} 

\subsection{Experimental Setup}\label{ch0-Planning-cp0}

\begin{figure}
    \centering
    \includegraphics[width=0.9\linewidth]{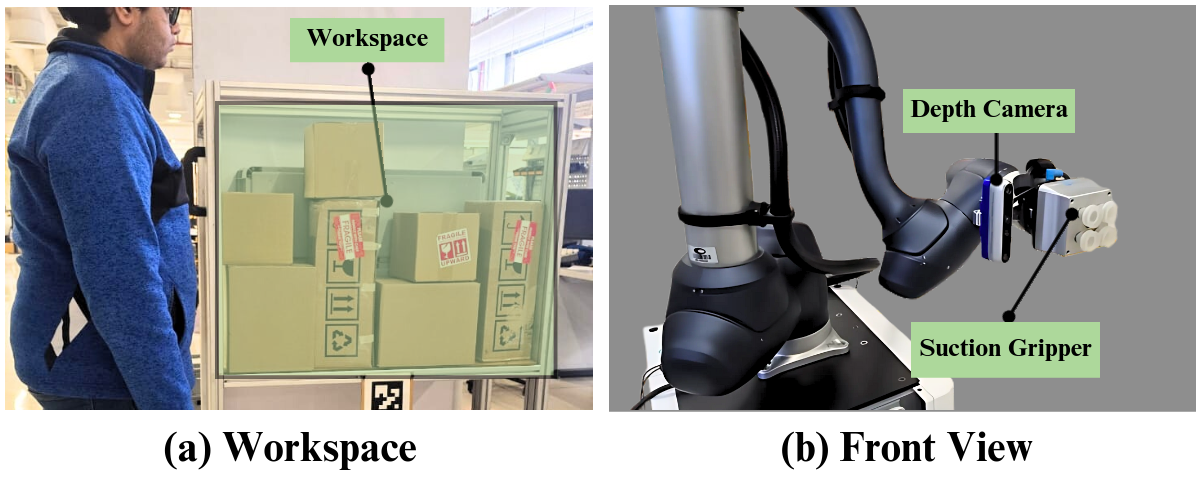}
    \caption{Experimental Setup}
    \label{fig:setup}
\end{figure}

Figure \ref{fig:setup} illustrates our experimental setup, which features a Doosan H2515 robotic arm with six degrees of freedom (6DOF). This means the arm can move in several different ways, making it very flexible. It is fitted with a suction gripper that helps it handle boxes precisely. Additionally, an Intel RealSense D455 depth camera is used to capture real-time spatial data, so the robot understands its surroundings.

A shelf measuring 100 cm by 30 cm by 160 cm was placed 104 cm away from the robot. This shelf served as the stacking area for our tests, which included picking a single box and clearing the shelf entirely. Additionally, we demonstrated the robustness of our approach using cardboard boxes of mixed sizes. Specifically, we used the following three sizes: 23 cm × 31 cm × 25 cm, 20 cm × 20 cm × 20 cm, and 50 cm × 17 cm × 17 cm. 

\subsection{Graphical User Interface}\label{ch0-Planning-cp0}
All components discussed in the methodology section are integrated into a graphical user interface, designed to enable effective human-robot collaboration through clear and transparent communication. Built with Flask, HTML, and CSS, the interface features a live camera feed with object detection overlays, alongside a chat-based interface that displays the robot's reasoning. Qwen2.5-7B-Instruct \cite{qwen2.5} is used as the LLM backend for Chain-of-Thought reasoning and decision-making, with Whisper and gTTS used for speech-to-text and text-to-speech, respectively, ensuring seamless interaction and fostering operator trust.
\subsection{Training Settings}\label{ch0-Planning-cp0}

Object segmentation for cardboard box detection is performed using a YOLOv11n-seg model, fine-tuned on the 8,401-image Online Stacked Cardboard Boxes Dataset (80\% training, 10\% validation, 10\% test). To improve performance across varying box rotations, training data was augmented with random rotations (-90° to 90°). Training, conducted on a system equipped with an RTX 3070 Ti (8GB VRAM), Intel i7-11700K processor, and 64GB RAM, achieved a test mAP of 0.87, with consistent validation performance. Separately, a YOLOv11n-seg model was fine-tuned for pointing recognition using a custom dataset of 156 manually labeled images with similar training settings as the previous model, yielding a mAP of 0.82. This pointing model provides initial segmentation masks for user interaction.



\subsection{Results}\label{ch0-Planning-cp0}

\begin{figure}[h!]
    \centering
    \includegraphics[width=0.75\linewidth]{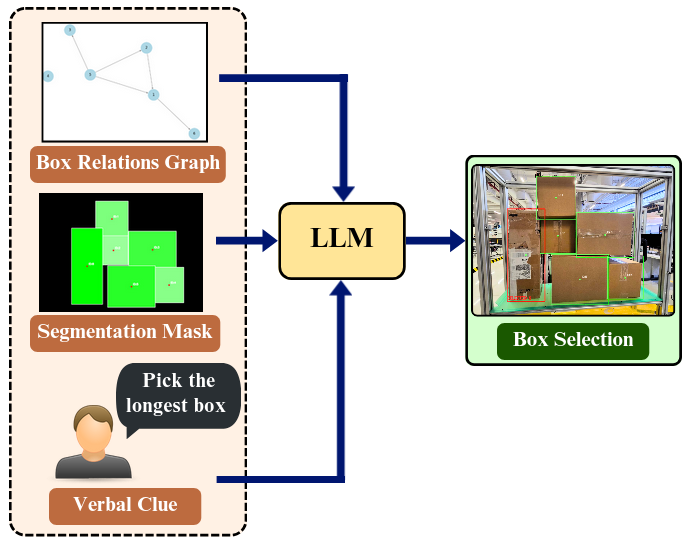}
    \caption{Overview of box selection based on verbal clues}
    \label{fig:verbal}
\end{figure}

Firstly, to assess the system's accuracy in interpreting human interaction, we conducted 10 trials using shelf images with varying box configurations. In each trial, a participant was asked to select a specific box first by pointing, and then by providing a verbal description. The system's box selection was recorded for both input modalities. The results of these experiments can be seen in table \ref{tab:pointing_recognition} and table \ref{tab:verbal_interpretation} and an overview of the pointing recognition based box selection and verbal clue based box selection can be seen in figures \ref{fig:graspplanner} and \ref{fig:verbal}.

\begin{figure*}[t!]
    \centering
    \includegraphics[width=0.8\linewidth]{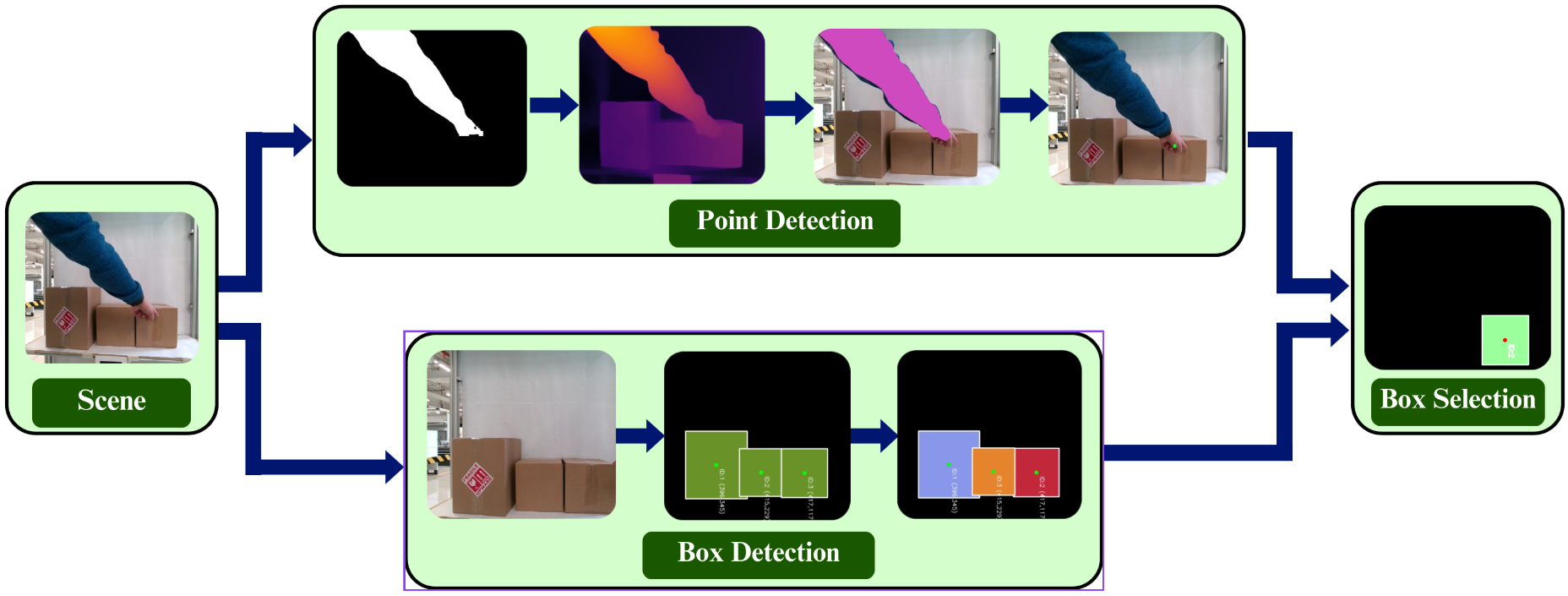}
    \caption{Overview of box selection based on pointing gesture}
    \label{fig:graspplanner}

\end{figure*}

\begin{table}[h!]
\centering
\caption{Pointing Gesture Recognition Results}
\label{tab:pointing_recognition}
\begin{tabular}{|c|c|c|c|}
\hline
Exp. No. & Ground Truth Box ID &  Selected Box ID & Success \\
\hline
1  & 3 & 3 & \ding{51} \\
2  & 2 & 2 & \ding{51} \\
3  & 5 & 5 & \ding{51} \\
4  & 4 & 4 & \ding{51} \\
5  & 1 & 2 & \ding{55} \\
\hline
\end{tabular}
\end{table}

\begin{table}[h!]
\centering
\caption{Verbal Clue Interpretation Results}
\label{tab:verbal_interpretation}
\begin{tabular}{|c|c|c|c|c|}
\hline
Exp. No. & Verbal Clue & Ground Truth ID & Selected ID & Success \\
\hline
1 & Third from left & 3 & 3 & \ding{51} \\
2 & Small blue & 2 & 4 & \ding{55} \\
3 & Box with label & 3 & 5 & \ding{55} \\
4 & Box near top & 4 & 4 & \ding{51} \\
5 & Box in middle & 1 & 1 & \ding{51} \\
\hline
\end{tabular}
\end{table}

  \begin{figure}[b!]
        \centering
        \includegraphics[width=0.85\linewidth]{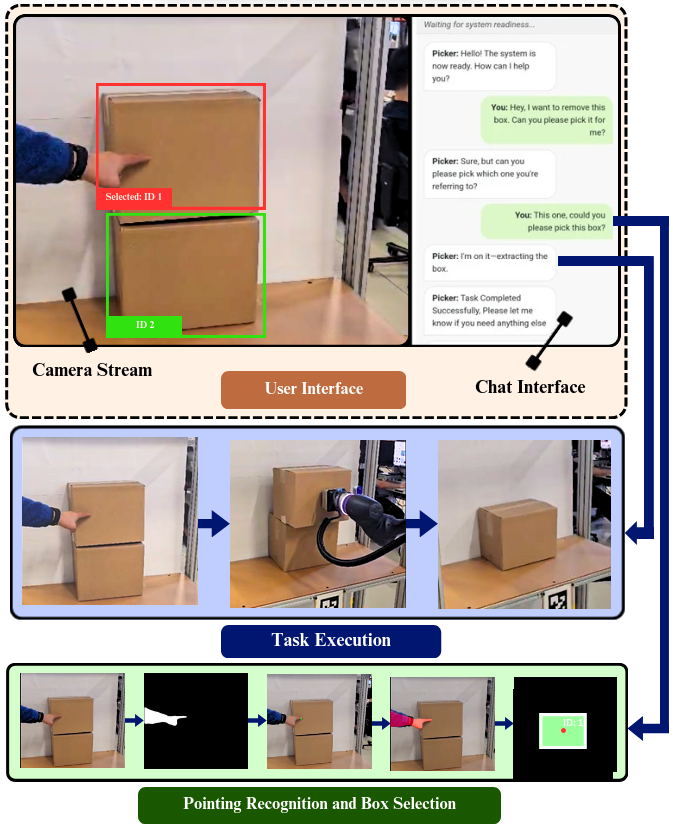}
        \caption{Overview of the gesture-guided box extraction scenario}
        \label{fig:box_extraction}
    \end{figure}

Secondly, to validate our proposed Human-Robot Collaboration Framework, we conducted experiments covering three different scenarios:
    \begin{enumerate}
    \item \textbf{Gesture-Guided Box Extraction:}
    In this simplest scenario, a human indicates a target box—often by pointing—and the robot uses its perception module to interpret the scene. This enables the robot to determine a safe extraction sequence without additional input. And overview of this can be seen in Figure \ref{fig:box_extraction}    
    \vspace{1em}
    \item \textbf{Collaborative Shelf Clearing:  }
    This scenario illustrates the benefits of shared tasks between the human and the robot. During a shelf clearance task, the robot and human work together by dividing the workload. For comparison, we also conducted experiments where the robot cleared the shelf alone and where the human performed the task independently. Results showed that human-robot collaboration consistently improved efficiency and safety. The results for this are shown in table \ref{tab:execution_time_comparison}. It is important to note, that human and robot collaboration is slower than human-only due to the operational speed of the manipulator arm being a bottleneck. An overview of this experiment is illustrated in Figure \ref{fig:Overview of the collaborative shelf clearning scenario} 

\begin{figure*}[t!]
    \centering
    \includegraphics[width=0.9\linewidth]{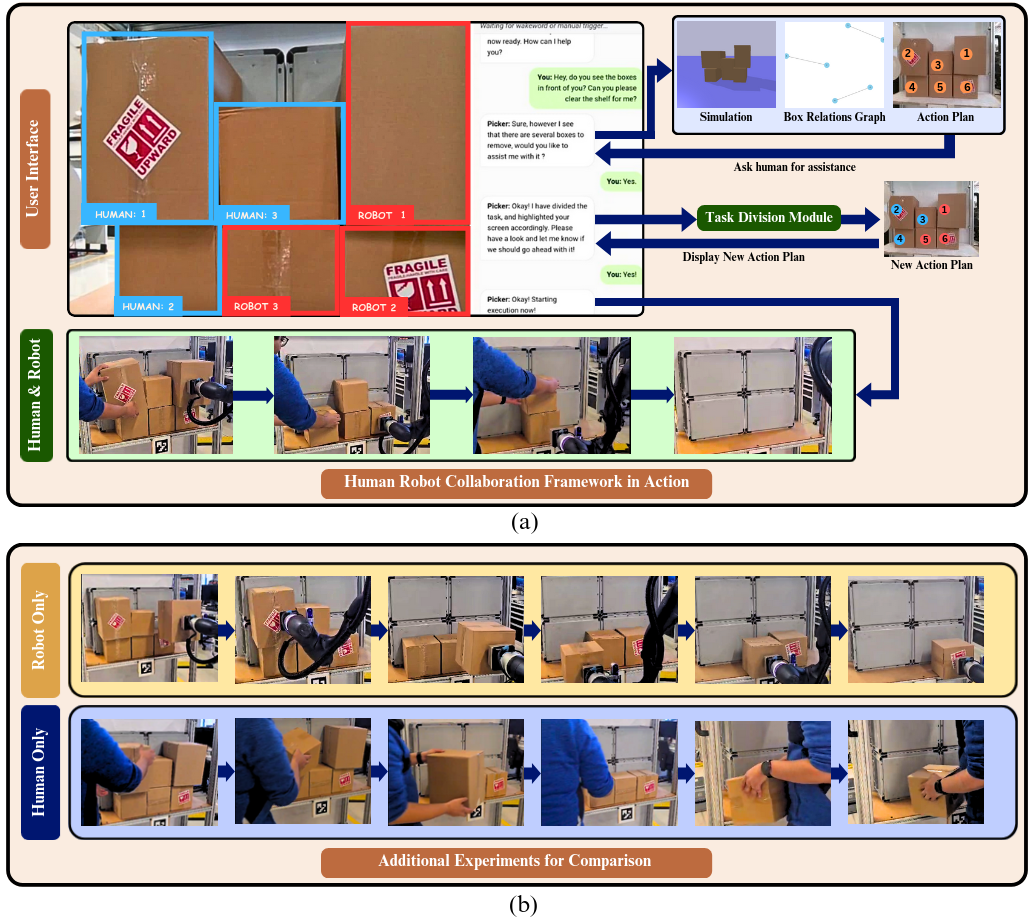}
    \caption{Overview for the collaborative shelf clearing scenario}
    \label{fig:Overview of the collaborative shelf clearning scenario}
\end{figure*}
\begin{table}[h!]
    \centering
    \caption{Execution Times (seconds) by Scenario}
    \label{tab:execution_time_comparison}
    \begin{tabular}{|c|c|c|}
        \hline
        Scenario & Execution Time (s) \\
        \hline
        Robot Only & 89.13 \\
        Human Only & 39.89 \\
        Human \& Robot & 42.21 \\
        \hline
    \end{tabular}
\end{table}

    \vspace{1em}
    \item \textbf{Collaborative Stability Assistance:}
   
    In this mode, the robot assists the human during the extraction process. For instance, while the human removes a target box, the robot either stabilizes the neighboring boxes (as seen in Figure \ref{fig:assit}) that are at risk of collapse or, alternatively, extracts the box while the human supports adjacent boxes (as seen in Figure \ref{fig:assit2}).
    
    \begin{figure}[h!]
        \centering
        \includegraphics[width=1\linewidth]{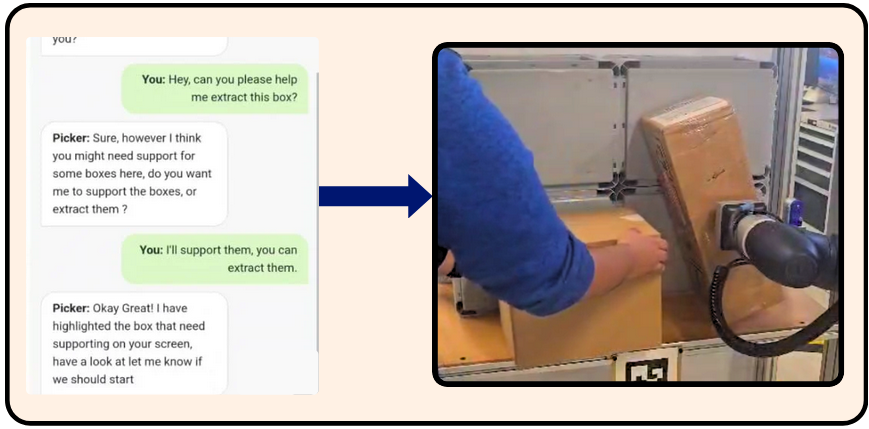}
        \caption{Human extracting box as the robot provides assistance}
        \label{fig:assit}
    \end{figure}

    \begin{figure}[h!]
            \centering
            \includegraphics[width=1\linewidth]{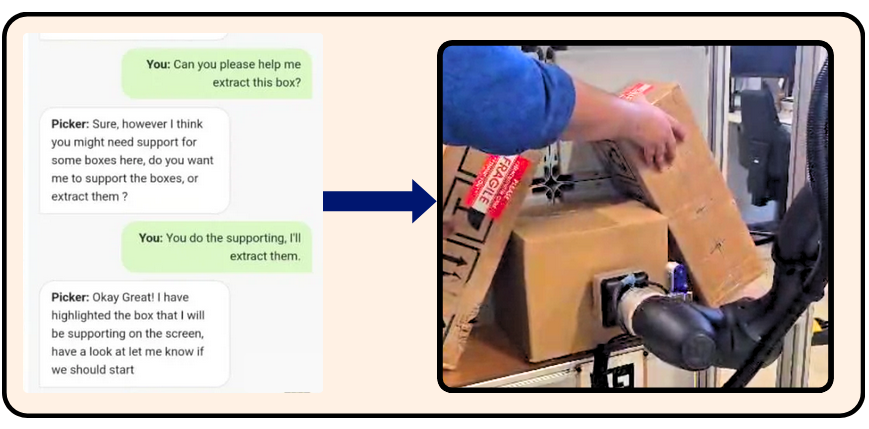}
            \caption{Robot extracting box as the human provides assistance}
            \label{fig:assit2}
    \end{figure}
    \end{enumerate}

\subsection{Discussion} \label{dis}

The experiments demonstrate that the Human-Robot Collaboration Framework enhances shelf-picking tasks. In the Single Box Picking scenario, the system identifies and extracts a target box by accurately interpreting the scene and planning a safe extraction sequence. In collaborative shelf clearing, human and robot share the task, resulting in increased efficiency (up to 2 times faster execution time) and safety compared to individual efforts. Furthermore, the collaborative support assistance demonstrates the framework's ability to facilitate in-contact collaboration, where the robot either stabilizes neighboring boxes or works with the human to maintain shelf integrity. The framework's reliance on visual cues and voice commands establishes clear, real-time communication, enabling effective decision-making in real-world settings.

\section{Conclusions}\label{conc}

In this paper, we proposed a novel multi-modal interaction framework, integrating a physics-informed Box Relations Graph (BRG) and an LLM-driven interface, to facilitate intuitive and safe shelf-picking operations. Real-world experiments validated the framework's effectiveness in gesture-guided extraction, collaborative shelf clearing, and stability assistance, demonstrating its practical application in human-centric warehouse environments. 

We plan on extending our work in the future by expanding the system's applicability to generalized robotic manipulation scenarios, implement it on mobile manipulator platforms, conduct more extensive experimental analyses, increase inference speed for real-time performance and explore advanced learning techniques to further refine the LLM’s reasoning and decision-making capabilities, aiming to create more adaptable and efficient human-robot teamwork beyond warehouse-specific tasks.

\bibliographystyle{IEEEtran}
\bibliography{references}  

\end{document}